\documentclass[10pt,twocolumn,letterpaper]{article}

\usepackage[pagenumbers]{cvpr} 

\usepackage{microtype}

\definecolor{cvprblue}{rgb}{0.21,0.49,0.74}
\usepackage[pagebackref,breaklinks,colorlinks,allcolors=cvprblue]{hyperref}
\usepackage{subcaption}
\usepackage{placeins}

\def\confName{CVPR}
\def\confYear{2027}

\title{CRA: Continuous Risk Assessment for Safety-critical Visual Navigation}

\author{
Lihao Zheng \quad Ying Wu \quad Chen Jiang \quad Randy Freeman\\[5pt]
Northwestern University\\[5pt]
{\ttfamily\small
\shortstack{
lihaozheng2027@u.northwestern.edu \quad yingwu@northwestern.edu\\
chenj@u.northwestern.edu \quad freeman@northwestern.edu
}
}
}

\begin{document}
\maketitle
\begin{abstract}

Traditional safety-critical control methods often rely on predefined constraints that separate states into safe and unsafe regions. While recent studies have extended geometric safety toward semantic safety, real-world safety is rarely purely binary. The risk associated with an object or human can vary continuously with its physical properties, like motion, and interaction context; for example, a person carrying a heavy object or a hot object moving at high speed may require a different safety response from an ordinary static obstacle. In this paper, we propose a vision-based continuous risk assessment framework for safety-critical visual navigation. Our method uses RGB and LiDAR observations to estimate four risk
factors (kinetic, potential, contact-severity, and thermal) and combines them into a continuous risk score for each assessed object or person. These estimates are integrated into a Poisson safety function with a Laplace guidance field as a safety filter, enabling online monitoring and intervention in open-world robot navigation. We demonstrate the proposed method in a simulated restaurant scene. Demonstration videos are available on our
\href{https://isaac-restaurant-cra.cra-demos.workers.dev}{project webpage}.

\end{abstract}    
\section{Introduction}
\label{sec:intro}

Navigation is one of the most important capabilities of autonomous systems and has been studied for decades~\cite{robotmotion_choset2005principles,navigationbook_latombe2012robot}. In navigation studies, vision plays an important role in localization, map construction, environment monitoring, and risk detection. With the development of camera-based perception, visual navigation has been widely studied in applications such as mobile robots~\cite{visual_navigation_survey_bonin2008}, autonomous driving~\cite{end_autonomous_driving_bojarski2016}, and aerial robots~\cite{UAV_visual_survey_lu2018}. Beyond geometric obstacle avoidance, recent research has shown that navigation agents should perceive objects, humans, and scene context for better decision-making~\cite{humblot2022navigation,chaplot2020object,savva2019habitat, krantz2020beyond}. This illustrates that scene understanding is a nontrivial component of visual navigation, especially in open-world robotic navigation. An object with the same geometry may have different contextual meanings; for example, a static object on the floor and a sharp or fast-moving object may require different navigation actions even if they occupy similar space. 

\begin{figure}[!t]
  \centering
\includegraphics[width=\columnwidth]{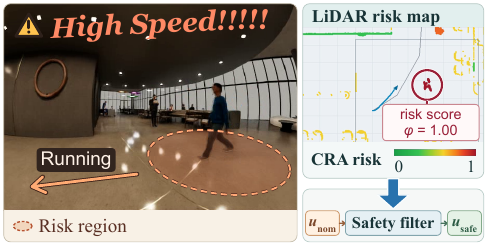}
  \caption{A running child approaches the robot. The robot ``sees'' the danger, and its controller responds immediately through the safety filter.}
  \label{fig:runner-teaser}
\end{figure}

Despite significant progress in visual navigation across a wide range of applications, navigation systems typically operate by producing a nominal trajectory or control command. However, when deployed in complex real-world environments, this nominal motion may become unsafe due to perception errors, occlusions, model mismatch, dynamic obstacles, or imperfect trajectory tracking~\cite{fastrack_herbert2017}. Therefore, safety should not only be considered inside the motion planner, but also enforced by an additional real-time safety layer before the robot executes its action. Safety filters provide such a mechanism by monitoring the proposed control input at runtime and intervening when necessary to ensure safety during execution~\cite{safety_filter_survey_hsu2023}.

Representative safety-filtering methods include control barrier functions (CBFs)~\cite{ames2019control}, model-predictive control (MPC)~\cite{zhang2020optimization}, Hamilton-Jacobi (HJ) reachability~\cite{bansal2017hamilton}, and data-driven safety filters~\cite{datadriven_sruvey_dawson2023safe}. These methods provide rigorous guarantees by maintaining the system within a prescribed safe set. The safe sets of these methods are often designed through predefined constraints~\cite{CBFQP_ames2016control, wabersich2021predictive}, learned from data~\cite{qin2021learning}, or constructed from occupancy maps and geometric primitives~\cite{molnar2023composing}. As a result, classical safety-filtering approaches are often ``semantically blind''~\cite{SAFESAGE2026}, focusing on geometric rather than contextual notions of safety: they can reason about where an obstacle is, but not necessarily what the obstacle is or why it may impose different levels of risk.

Recent work on semantic safety addresses this limitation across several robotic settings, including robot manipulation~\cite{brunke2025semantically}, open-world robot navigation and embodied AI~\cite{sermanet2025generating}, and autonomous driving~\cite{greer2026vision}. Context-aware robotic safety is especially important in open-world environments~\cite{why_semantic_safe_robey2026}, where the same action may be safe or unsafe depending on surrounding humans, objects, and physical context. Although semantic reasoning has improved safety-critical robot decision-making, many existing approaches operate primarily at the planning or task-reasoning level, where LLMs and vision-language models (VLMs) are used to ground language instructions and generate semantic safety rules~\cite{ahn2022can,sermanet2025generating,zhan2025sentinel}. However, these methods are often designed for high-level planning rather than real-time safety filtering. Directly using LLMs or VLMs inside the safety filter loop can introduce high latency, while the safety filter and controller require low-latency monitoring and intervention~\cite{visual_navigation_survey_bonin2008,ravichandran2026contextual}. Moreover, existing semantic safety methods often convert context into discrete rules or constraints, but real-world risk is continuous and context-dependent, varying with object properties, motion, and interaction context.

This continuous and context-dependent nature of risk suggests that safety cannot be completely captured by fixed discrete rules. Prior work on fuzzy safety analysis has similarly argued that safety is difficult to define exactly and may involve different degrees of hazard~\cite{gentile2003development}. In open-world visual navigation, such ambiguity naturally arises: a person walking freely and a person carrying a heavy object may not be strictly classifiable as simply safe or unsafe, even if they are at the same distance from the robot; instead, the robot should assess how safe the interaction is under the current context. Similarly, a static object and a sharp or fast-moving object may involve different levels of risk despite similar geometry. Therefore, rather than relying only on predefined semantic rules, a robot should assess how safe the current scene is and use this continuous risk assessment to guide real-time safety filtering.


Motivated by this observation, we propose a vision-based continuous risk assessment framework for safety-critical visual navigation. Instead of assigning predefined safe or unsafe labels to semantic objects, our method analyzes scene risk through four physical factors derived from visual reasoning, including object properties, motion, and interaction context. We integrate continuous risk assessment into a safety filter based
on Safe-SAGE~\cite{SAFESAGE2026}. The risk scores set the normal
boundary flux of the Laplace guidance field
(LGF)~\cite{PSF_LGF_bahati2026risk}, which shapes the Poisson safety
function (PSF)~\cite{PSF_bahati2025dynamic}. The filter uses both
fields to adjust the robot's commands online through
MPC and CBF filtering~\cite{bena2025geometry}.
\section{Related Work}

\subsection{Geometric Safety Filters}

Geometric safety filters constrain nominal control inputs to keep the system within a prescribed safe set~\cite{ames2019control,bansal2017hamilton,wabersich2021predictive}. In robot navigation, these safe sets are often constructed from geometric representations, such as pairwise safety distances~\cite{wang2017safety}, point-cloud-based obstacle representations~\cite{de2024point,dai2024sailing}, and occupancy-grid maps~\cite{raja2024safe}. These methods provide rigorous runtime safety guarantees, but they usually do not distinguish objects by their semantic or physical risk. Our work extends geometric safety filtering by using visual reasoning to construct continuous, \textit{context-aware} risk assessments for the safety layer.

\subsection{Semantic Safety}

Semantic safety utilizes language, vision-language models, and scene understanding for robotic safety. Language-conditioned planning grounds semantic instructions to robotic affordances~\cite{ahn2022can}. Other works study semantic safety benchmarks for evaluating unsafe robot behaviors~\cite{sermanet2025generating}. With the advancement of LLMs and VLMs, recent safety-filter methods further connect semantic information to control-level safety through CBFs~\cite{brunke2025semantically,ravichandran2026contextual} or HJ reachability~\cite{santos2025updating}. Other works use latent-space safe sets to address hard-to-model safety constraints and out-of-distribution (OOD) failures~\cite{seo2025uncertainty,nakamura2025train}. However, foundation-model reasoning can introduce latency for real-time control~\cite{VLASafe_li2026vision}. Safe-SAGE~\cite{SAFESAGE2026} provides a semantic-aware framework for real-time safety filtering by modulating PSF/LGF-based safety fields with visual semantics. However, these methods often represent semantic context as rules, object classes, or constraints. Our work builds on this framework but estimates continuous risk from visual reasoning for real-time safety filtering.

\subsection{Risk-aware and Robust Safety}

Risk-aware and robust safety methods address uncertainty in safety-critical control, including measurement and perception uncertainty~\cite{dean2021guaranteeing,cosner2021measurement}, model uncertainty and disturbances~\cite{dacs2025robust, Jason_uncertain}, stochastic uncertainty~\cite{clark2021control,cosner2023robust}, and CVaR-based risk-aware formulations~\cite{ahmadi2021risk}. These methods improve safety when the robot state, dynamics, measurements, or disturbances are uncertain. However, they usually assume a predefined safety specification, such as a geometric distance constraint. Our work represents safety on a continuous scale by estimating how
the risk posed by an object or interaction varies with object
properties, relative motion, and the robot's planned route.

\subsection{Fuzzy Safety}

Fuzzy views of safety have been explored more in other safety-critical domains than in autonomous robotic systems. Prior work on inherent safety argues that safety is difficult to define exactly and should be evaluated by degrees rather than only by strict safe/unsafe categories~\cite{gentile2003development}. Fuzzy logic has also been used for qualitative safety modeling in marine systems~\cite{sii2001fuzzy} and for traffic safety to avoid crisp safety thresholds~\cite{mattas2020fuzzy}. These works show that safety can be better represented as a gradual concept, such as ``not very safe'' or ``highly unsafe,'' rather than only as a strict safe/unsafe category. However, this perspective has not been widely explored in semantic safety for autonomous visual navigation. We apply this view to visual navigation by representing risk
on a continuous scale.

\subsection{Visual Physical Reasoning}

Visual physical reasoning studies how to infer object properties,
dynamics, and interactions from images and videos. Prior work
examines object motion and interaction dynamics~\cite{watters2017visual},
collision events~\cite{yi2019clevrer}, physical dynamics such as rolling,
sliding, and falling~\cite{bear2021physion}, and block-tower
stability~\cite{lerer2016learning}. Other work recognizes
materials~\cite{bell2015material} and affordances~\cite{do2018affordancenet}. In our work, we combine RGB and LiDAR observations to estimate kinetic, potential,
contact-severity, and thermal risk for visual navigation.

\section{Background}
\label{sec:background}

\subsection{Safety Filters}

\textbf{Safe Sets.}
Let $\mathbf{q}\in\mathbb{R}^{n_q}$ denote the state and
$\mathbf{u}\in\mathbb{R}^{m_q}$ the control input, with admissible
inputs in $\mathcal{U}$. A continuously differentiable function
$h:\mathbb{R}^{n_q}\to\mathbb{R}$ defines the \textit{safe set} as
\begingroup
\setlength{\abovedisplayskip}{5pt}
\setlength{\belowdisplayskip}{5pt}
\setlength{\abovedisplayshortskip}{3pt}
\setlength{\belowdisplayshortskip}{3pt}
\begin{equation}
    \mathcal{C} := \{\mathbf{q}\in\mathbb{R}^{n_q}\mid h(\mathbf{q})\geq 0\},
\end{equation}
\endgroup
with boundary and interior
\begingroup
\setlength{\abovedisplayskip}{5pt}
\setlength{\belowdisplayskip}{5pt}
\setlength{\abovedisplayshortskip}{3pt}
\setlength{\belowdisplayshortskip}{3pt}
\begin{equation}
\begin{aligned}
    \partial\mathcal{C}
    &:= \{\mathbf{q}\in\mathbb{R}^{n_q}\mid h(\mathbf{q})=0\}, \\
    \mathrm{int}(\mathcal{C})
    &:= \{\mathbf{q}\in\mathbb{R}^{n_q}\mid h(\mathbf{q})>0\}.
\end{aligned}
\end{equation}
\endgroup
The role of a safety filter is to render $\mathcal{C}$ forward invariant,
ensuring that trajectories initialized in $\mathcal{C}$ remain in
$\mathcal{C}$ for all future time.

\textbf{Control Barrier Functions (CBFs)}~\cite{ames2019control} provide a sufficient condition for
forward invariance of the \textit{safe set}. For control-affine dynamics
$\dot{\mathbf{q}}=\mathbf{f}(\mathbf{q})+\mathbf{g}(\mathbf{q})\mathbf{u}$,
a continuously differentiable function $h$ satisfying
$\nabla h(\mathbf{q})\neq 0$ when $h(\mathbf{q})=0$ is a CBF on
$\mathcal{C}$ if there exists $\gamma\in\mathcal{K}_{\infty}^{e}$
such that, for all $\mathbf{q}\in\mathbb{R}^{n_q}$,
\begingroup
\setlength{\abovedisplayskip}{5pt}
\setlength{\belowdisplayskip}{5pt}
\setlength{\abovedisplayshortskip}{3pt}
\setlength{\belowdisplayshortskip}{3pt}
\begin{equation}
    \sup_{\mathbf{u}\in\mathbb{R}^{m_q}}
    \left[
    L_{\mathbf{f}}h(\mathbf{q})+
    L_{\mathbf{g}}h(\mathbf{q})\mathbf{u}
    \right]
    \geq
    -\gamma(h(\mathbf{q})).
\end{equation}
\endgroup
Choosing control inputs that satisfy
\begingroup
\setlength{\abovedisplayskip}{5pt}
\setlength{\belowdisplayskip}{5pt}
\setlength{\abovedisplayshortskip}{3pt}
\setlength{\belowdisplayshortskip}{3pt}
\begin{equation}
    L_{\mathbf{f}}h(\mathbf{q})+
    L_{\mathbf{g}}h(\mathbf{q})\mathbf{u}
    \geq
    -\gamma(h(\mathbf{q}))
\end{equation}
\endgroup
renders $\mathcal{C}$ forward invariant.

Let $\mathbf{F}(\mathbf{q}_k,\mathbf{u}_k)$ denote the state one
step ahead. The discrete-time CBF (DTCBF)
condition~\cite{agrawal2017discrete} can be
\begingroup
\setlength{\abovedisplayskip}{5pt}
\setlength{\belowdisplayskip}{5pt}
\setlength{\abovedisplayshortskip}{3pt}
\setlength{\belowdisplayshortskip}{3pt}
\begin{equation}
    h(\mathbf{F}(\mathbf{q}_k,\mathbf{u}_k)) \geq \rho h(\mathbf{q}_k),
    \quad \rho\in[0,1].
\end{equation}
\endgroup

\textbf{Safety-Critical Control}~\cite{CBFQP_ames2016control}
uses CBF constraints to minimally modify a nominal input while enforcing
safety. In the real-time case, the safety filter solves a one-step projection
problem. Given a nominal control input $\mathbf{u}_{\mathrm{nom}}$, the
continuous-time CBF safety filter is formulated as
\begingroup
\setlength{\abovedisplayskip}{5pt}
\setlength{\belowdisplayskip}{5pt}
\setlength{\abovedisplayshortskip}{3pt}
\setlength{\belowdisplayshortskip}{3pt}
\begin{equation}
\begin{aligned}
&\mathbf{u}_{\mathrm{safe}}
 = \operatorname*{arg\,min}_{\mathbf{u}\in\mathbb{R}^{m_q}}\;
   \lVert\mathbf{u}-\mathbf{u}_{\mathrm{nom}}\rVert_2^2 \\
&\text{s.t.}\;
 L_{\mathbf{f}}h(\mathbf{q})
 + L_{\mathbf{g}}h(\mathbf{q})\mathbf{u}
 \geq -\gamma(h(\mathbf{q})).
\end{aligned}
\label{eq:cbf_qp}
\end{equation}
\endgroup

Predictive safety filters extend this idea over a finite horizon by optimizing
a sequence of inputs while enforcing safety constraints along the predicted
trajectory~\cite{borrelli2017predictive,bena2025geometry}. A CBF-constrained
MPC safety filter can be written as
\begingroup
\setlength{\abovedisplayskip}{5pt}
\setlength{\belowdisplayskip}{5pt}
\setlength{\abovedisplayshortskip}{3pt}
\setlength{\belowdisplayshortskip}{3pt}
\begin{equation}
\begin{aligned}
&\mathbf{u}^{\star}_{0:N-1}
 = \operatorname*{arg\,min}_{\mathbf{u}_{0:N-1}}\;
   \sum_{k=0}^{N-1}
   \lVert\mathbf{u}_k-\mathbf{u}_{\mathrm{nom},k}\rVert_2^2 \\
&\text{s.t.}\;
\begin{aligned}[t]
&\mathbf{q}_{k+1} = \mathbf{F}(\mathbf{q}_k,\mathbf{u}_k), \\
&h(\mathbf{q}_{k+1}) \geq \rho\,h(\mathbf{q}_k), \\
&\mathbf{u}_k\in\mathcal{U}, \quad k=0,\ldots,N-1.
\end{aligned}
\end{aligned}
\label{eq:mpc_safety_filter}
\end{equation}
\endgroup

\subsection{Poisson Safety Functions and Laplace Guidance Fields}
\label{sec:psf_lgf}

\textbf{Poisson Safety Functions (PSFs)}~\cite{PSF_bahati2025dynamic}
are functions that provide a continuous representation of spatial safety
directly from environmental occupancy data. Let $\mathbf{y}=(x,y,z)\in\mathbb{R}^{3}$ denote the robot's \textit{spatial
coordinates}, and let $\Omega\subset\mathbb{R}^{3}$ be a smooth, open, bounded,
and connected set representing free space, with $\partial\Omega$ denoting the
surfaces of occupied regions. Given a smooth negative forcing function
$\hat{f}:\Omega\rightarrow\mathbb{R}_{<0}$, a PSF $h_0$ is obtained by solving
the Dirichlet boundary-value problem
\begingroup
\setlength{\abovedisplayskip}{5pt}
\setlength{\belowdisplayskip}{5pt}
\setlength{\abovedisplayshortskip}{3pt}
\setlength{\belowdisplayshortskip}{3pt}
\begin{equation}
\begin{cases}
    \Delta h_0(\mathbf{y}) = \hat{f}(\mathbf{y}),
    & \mathbf{y}\in\Omega, \\
    h_0(\mathbf{y}) = 0,
    & \mathbf{y}\in\partial\Omega,
\end{cases}
\label{eq:psf}
\end{equation}
\endgroup
where
\(
    \Delta
    =
    \frac{\partial^2}{\partial x^2}
    +
    \frac{\partial^2}{\partial y^2}
    +
    \frac{\partial^2}{\partial z^2}
\)
denotes the Laplace operator.

Under appropriate regularity assumptions on $\Omega$, a smooth forcing
function $\hat{f}\in C^{\infty}(\Omega)$ yields a smooth solution
$h_0\in C^{\infty}(\Omega)$ to~\eqref{eq:psf}~\cite{gilbarg1977elliptic}.
As demonstrated in~\cite{PSF_bahati2025dynamic}, this solution characterizes
the safe set $\mathcal{C}$ such that
$\Omega=\mathrm{int}(\mathcal{C})$ and
$\partial\mathcal{C}=\partial\Omega$, and can be used to construct
safety filters under appropriate assumptions on the system dynamics.

\textbf{Laplace Guidance Fields (LGFs)}~\cite{PSF_LGF_bahati2026risk}
are vector fields that encode safe motion directions through prescribed
boundary fluxes along occupied-region boundaries. Let
$\hat{\mathbf{n}}:\partial\Omega\rightarrow\mathbb{R}^{3}$ denote the unit
normal pointing outward from the free space $\Omega$ toward occupied regions,
and let $b:\partial\Omega\rightarrow\mathbb{R}_{<0}$ denote a prescribed
boundary flux. A guidance field
$\mathbf{v}:\Omega\rightarrow\mathbb{R}^{3}$ satisfies
\begingroup
\setlength{\abovedisplayskip}{5pt}
\setlength{\belowdisplayskip}{5pt}
\setlength{\abovedisplayshortskip}{3pt}
\setlength{\belowdisplayshortskip}{3pt}
\begin{equation}
    \mathbf{v}(\mathbf{y})
    =
    b(\mathbf{y})\hat{\mathbf{n}}(\mathbf{y}),
    \qquad
    \mathbf{y}\in\partial\Omega.
\label{eq:guidance_boundary}
\end{equation}
\endgroup
The prescribed boundary values are extended throughout $\Omega$ by solving a
vector Laplace Dirichlet problem. For
$\mathbf{v}=(v_x,v_y,v_z)$, each component satisfies
\begingroup
\setlength{\abovedisplayskip}{5pt}
\setlength{\belowdisplayskip}{5pt}
\setlength{\abovedisplayshortskip}{3pt}
\setlength{\belowdisplayshortskip}{3pt}
\begin{equation}
\begin{cases}
    \Delta v_i(\mathbf{y})=0,
    & \mathbf{y}\in\Omega, \\
    v_i(\mathbf{y})
    =
    b(\mathbf{y})\hat{n}_i(\mathbf{y}),
    & \mathbf{y}\in\partial\Omega,
\end{cases}
\qquad i\in\{x,y,z\}.
\label{eq:lgf}
\end{equation}
\endgroup
Since $b(\mathbf{y})<0$, the guidance field points opposite to the outward
normal, away from occupied regions and toward the interior of the free space.

\textbf{Safety Constraints with PSFs and LGFs.}
Although an LGF $\mathbf{v}$ is generally non-conservative and need not satisfy
$\mathbf{v}\equiv\nabla h_0$ in $\Omega$, it satisfies
$\mathbf{v}\parallel\nabla h_0$ on $\partial\Omega$. As established
in~\cite{PSF_LGF_bahati2026risk,SAFESAGE2026}, this boundary alignment allows
the LGF to be combined with the PSF to enforce safety for first-order spatial
dynamics.

For the single-integrator model $\dot{\mathbf{y}}=\mathbf{u}$, a locally
Lipschitz feedback law $\mathbf{u}=\mathbf{k}(\mathbf{y})$ satisfying
\begin{equation}
    \mathbf{v}(\mathbf{y})^\top\mathbf{u}
    \geq
    -\gamma h_0(\mathbf{y}),
    \qquad
    \gamma>0,
    \quad
    \mathbf{y}\in\mathcal{C},
\label{eq:psf_lgf_constraint}
\end{equation}
renders the PSF-defined safe set $\mathcal{C}$ forward invariant.
\section{Method}

\subsection{System Overview}
\label{sec:system_overview}

\begin{figure*}[t]
    \centering
    \includegraphics[
        width=\textwidth,
        height=0.55\textheight,
        keepaspectratio
    ]{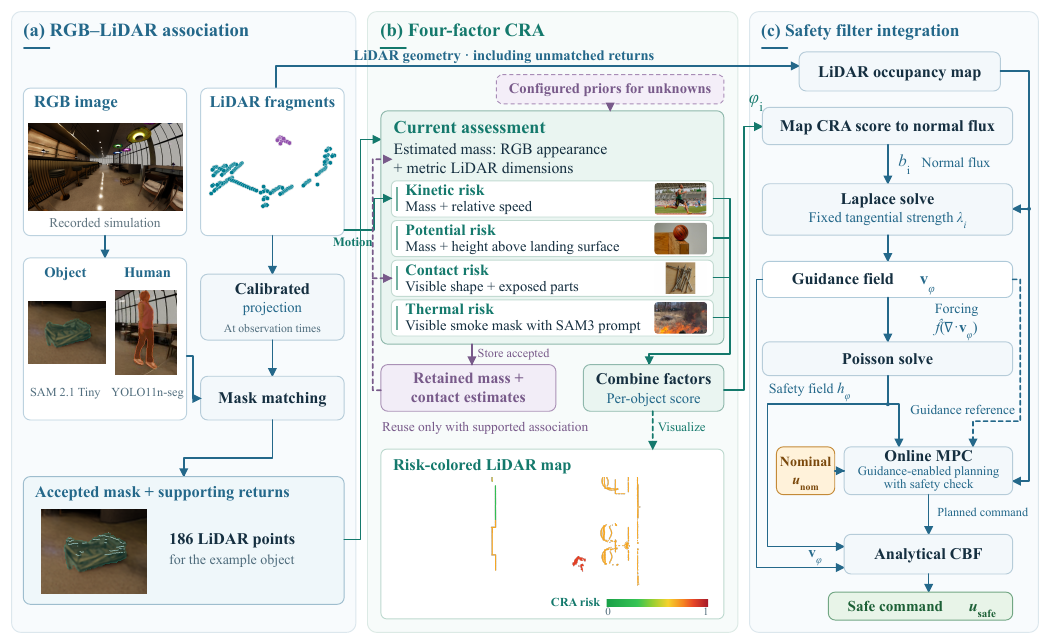}
    \caption{Overview of the CRA framework. RGB--LiDAR association
    provides object evidence for four-factor risk estimation. The
    resulting per-object risk scores condition Safe-SAGE navigation.}
    \label{fig:pipeline}
\end{figure*}

Figure~\ref{fig:pipeline} summarizes our framework, which connects
RGB--LiDAR perception to risk-conditioned navigation through continuous
risk assessment (CRA). For planning and safety filtering, we model
the robot as a planar unicycle. Let
$\mathbf{q}=(p_x,p_y,\psi)^\top\in\mathbb{R}^{2}\times\mathbb{S}^{1}$
denote the robot pose and
$\mathbf{u}=(v_{\mathrm{lin}},\omega)^\top\in\mathcal{U}\subset\mathbb{R}^{2}$
the control input. The model is given by
\begingroup
\setlength{\abovedisplayskip}{5pt}
\setlength{\belowdisplayskip}{5pt}
\setlength{\abovedisplayshortskip}{3pt}
\setlength{\belowdisplayshortskip}{3pt}
\begin{equation}
    \dot{\mathbf{q}}
    =
    \begin{bmatrix}
        \cos\psi & 0 \\
        \sin\psi & 0 \\
        0       & 1
    \end{bmatrix}
    \mathbf{u},
    \label{eq:navigation_dynamics}
\end{equation}
\endgroup
where $(p_x,p_y)$ is the planar position, $\psi$ is the heading,
$v_{\mathrm{lin}}$ and $\omega$ are the forward and angular velocities,
and $\mathcal{U}$ specifies the admissible velocity commands.

An obstacle-unaware waypoint controller generates the nominal input
$\mathbf{u}_{\mathrm{nom}}$ along a prescribed route. Our safety layer
filters this input using environmental geometry and CRA-derived risk
to obtain $\mathbf{u}_{\mathrm{safe}}$.

\subsection{RGB--LiDAR Association}
\label{sec:rgb_lidar_evidence}

At each update, we process an RGB image $I_{t_C}$ and LiDAR returns
$\mathcal{P}_{t_L}=\{\mathbf{p}_j\}$, retaining their source timestamps.
LiDAR clustering and tracking organize the observed returns into
geometric fragments. We use pretrained SAM 2.1 Tiny~\cite{ravi2025sam}
to propose class-agnostic instance masks $\{M_i\}$ and COCO-pretrained
YOLO11n-seg~\cite{Jocher_Ultralytics_YOLO26_Unified_2026,lin2014microsoft} to obtain
instance masks and class cues for humans.

We associate visual instances with observed 3D returns through calibrated
projection, as used in projection-based camera--LiDAR
fusion~\cite{vora2020pointpainting}. Using sensor calibration and the ego
poses at the corresponding observation times, we project each LiDAR
return as
\begingroup
\setlength{\abovedisplayskip}{5pt}
\setlength{\belowdisplayskip}{5pt}
\setlength{\abovedisplayshortskip}{3pt}
\setlength{\belowdisplayshortskip}{3pt}
\begin{equation}
    \mathbf{z}_j =
    \Pi_C\!\left(
        \mathbf{R}_{CL}\mathbf{p}_j+\mathbf{t}_{CL}
    \right),
    \label{eq:lidar_image_projection}
\end{equation}
\endgroup
where $(\mathbf{R}_{CL},\mathbf{t}_{CL})$ transforms the point from the
LiDAR frame at $t_L$ to the camera frame at $t_C$, and $\Pi_C$ is the
calibrated fisheye projection. The resulting $\mathbf{z}_j$ is the image
coordinate of return $\mathbf{p}_j$.

For each visible LiDAR fragment, we use a projected return as a point
prompt for SAM to obtain a candidate mask~\cite{xu2025sampro3d}. LiDAR may sometimes split a single static object into several fragments.
When multiple fragments consistently align with the same SAM mask,
we group their verified returns into one object assessment.

For each accepted match, we pass the SAM mask, the associated LiDAR
returns, their original observation times, and the SAM-derived mask
confidence to the physical-factor estimators. When the YOLO human
mask agrees with the associated SAM mask in the same RGB image,
we use human-specific priors for mass and body contact. LiDAR returns without an accepted image match remain in the occupancy
map.

\subsection{Four-Factor Risk Assessment}
\label{sec:four_factor_risk}

\begin{figure*}[!t]
    \centering
    \captionsetup[subfigure]{
        font=small,
        labelfont=bf,
        labelformat=parens,
        labelsep=space,
        justification=raggedright,
        singlelinecheck=false,
        skip=3pt
    }

    \begin{subfigure}[t]{0.485\textwidth}
        \vspace{0pt}
        \parbox[c][1.65in][c]{\linewidth}{%
            \centering
            \includegraphics[width=\linewidth,height=1.65in,keepaspectratio]
                {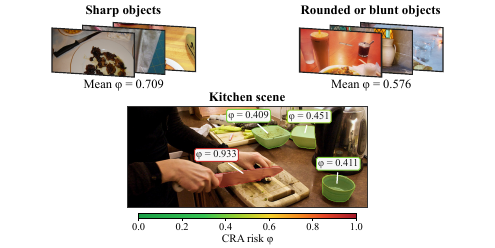}%
        }
        \caption{\textbf{Contact risk.} We evaluate 50 sharp objects and
        50 rounded or blunt objects from PACO~\cite{ramanathan2023paco}.
        CRA estimates the contact risk from the object masks predicted by YOLO~\cite{Jocher_Ultralytics_YOLO26_Unified_2026} while the other factor inputs are held fixed. Sharp objects
        receive a higher mean risk score ($\phi=0.709$) than rounded or
        blunt objects ($\phi=0.576$).}
        \label{fig:cra_contact}
    \end{subfigure}\hfill
    \begin{subfigure}[t]{0.485\textwidth}
        \vspace{0pt}
        \parbox[c][1.65in][c]{\linewidth}{%
            \centering
            \includegraphics[width=\linewidth,height=1.65in,keepaspectratio]
                {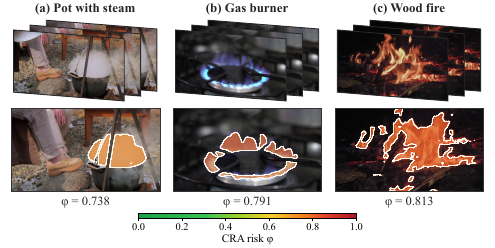}%
        }
        \caption{\textbf{Thermal risk.} We evaluate CRA on three RGB video
        clips showing a steaming pot, a gas burner, and a wood
        fire.\protect\footnotemark\ Each displayed $\phi$ is computed from
        the $0.5$~s temporal window ending at the pictured frame. As expected,
        the gas burner and wood fire receive higher risk scores than the
        steaming pot, which agrees with CRA's thermal-risk assessment.}
        \label{fig:cra_thermal}
    \end{subfigure}

    \par\vspace{1pt}

    \begin{subfigure}[t]{0.485\textwidth}
        \vspace{0pt}
        \parbox[c][1.80in][c]{\linewidth}{%
            \centering
            \includegraphics[width=\linewidth,height=1.80in,keepaspectratio]
                {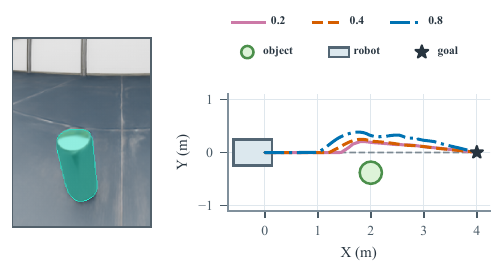}%
        }
        \caption{\textbf{Kinetic risk.} We evaluate CRA as the robot passes
        the same static object at requested speeds of $0.2$, $0.4$, and
        $0.8$~m/s. The mean CRA scores while passing the obstacle were
        $\phi=0.4489$, $0.4522$, and $0.4630$, respectively.
        Higher requested speeds produced larger detours in these runs.}
        \label{fig:cra_kinetic}
    \end{subfigure}\hfill
    \begin{subfigure}[t]{0.485\textwidth}
        \vspace{0pt}
        \parbox[c][1.80in][c]{\linewidth}{%
            \centering
            \includegraphics[width=\linewidth,height=1.80in,keepaspectratio]
                {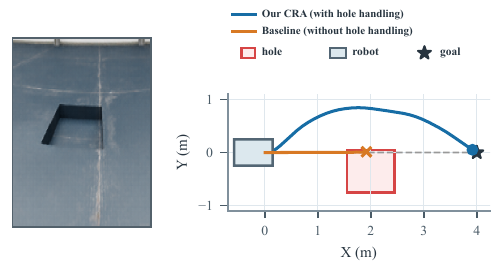}%
        }
        \caption{\textbf{Potential risk.} We tilt the LiDAR assembly
        $17^\circ$ downward to observe the bottom of a shallow hole and
        estimate its potential risk factor. CRA represents the
        hole as a risk region and assigns it a CRA risk score. With CRA, the robot goes around the hole and reaches the goal. The baseline Safe-SAGE~\cite{SAFESAGE2026} controller
        continues along the nominal route and the robot falls into the hole.}
        \label{fig:cra_dropoff}
    \end{subfigure}
    \caption{Continuous risk assessment results for the four physical
    risk factors.}
    \label{fig:cra_factors}
\end{figure*}

\begin{figure}[!t]
    \centering
    \includegraphics[width=0.75\columnwidth]{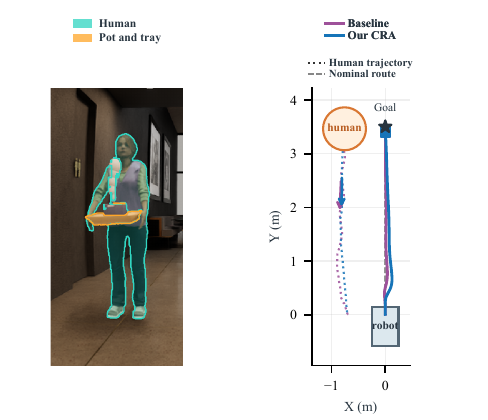}
    \caption{CRA navigation with a human carrying a
    steaming pot. As CRA 	assesses the person and pot together, the robot makes a larger measured detour compared with the baseline Safe-SAGE~\cite{SAFESAGE2026} controller with only human risk detected.}
    \label{fig:cra_combined}
\end{figure}

\begin{figure*}[!t]
    \centering
    \includegraphics[
        width=\textwidth,
        height=0.30\textheight,
        keepaspectratio
    ]{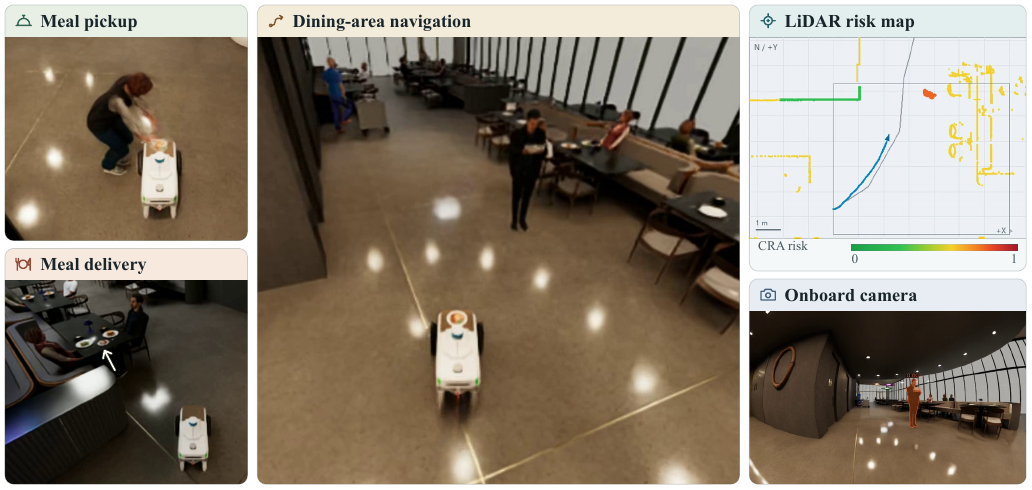}
    \caption{Restaurant service mission in simulation.}
    \label{fig:restaurant_navigation}
\end{figure*}

For each object or human $i$ with an accepted RGB--LiDAR association,
CRA estimates four risk factors: kinetic, potential, contact-severity,
and thermal risk. Each factor $f$ is represented by a normalized
severity $s_{if}\in[0,1]$, an interaction score $\ell_{if}\in[0,1]$,
and an evidence confidence $c_{if}\in[0,1]$.
Severity reflects how serious each type of risk is, as estimated
from physical quantities or visual proxies. The interaction score
reflects how likely the robot is to encounter that risk along its
planned route. Each factor input retains its source, observation
time, and uncertainty bounds. 
We use the upper bounds of accepted factor inputs to compute severity
and interaction, and denote these bounds with a superscript $+$.
We write $[x]_0^1=\min(1,\max(0,x))$ and $[x]_+=\max(0,x)$.

\paragraph{Mass estimation.}
Given a masked RGB crop $J_i$ and three LiDAR-derived dimensions
$\mathbf{d}_i$, we estimate object mass as
\begingroup
\setlength{\abovedisplayskip}{5pt}
\setlength{\belowdisplayskip}{5pt}
\setlength{\abovedisplayshortskip}{3pt}
\setlength{\belowdisplayshortskip}{3pt}
\begin{equation}
    \widehat{m}_i =
    \exp\!\left(
        g_{\theta}\!\left[
            E_{\mathrm{img}}(J_i),
            \operatorname{norm}(\log\mathbf{d}_i)
        \right]
    \right),
    \label{eq:visual_mass}
\end{equation}
\endgroup
where $E_{\mathrm{img}}$ is a frozen ImageNet-pretrained
MobileNetV3-Small encoder~\cite{howard2019searching,deng2009imagenet}.
The regression head $g_{\theta}$ combines the image features with the
normalized log dimensions. We train it on product images, dimensions,
and catalog weights from ABO~\cite{collins2022abo} using a Smooth-L1
loss on log mass. We keep each product in a single data split.
During navigation, CRA treats the predicted mass as a rough estimate and uses a preset margin to form an uncertainty range.

\paragraph{Route interaction.}

Interaction is obtained from a geometric route-overlap score.
Given the robot's current position and the target waypoint provided
by the navigation planner, let $d_i$ be the shortest planar distance
between the robot-to-waypoint segment and the object's path predicted
over $T_{\mathrm{int}}$ seconds at its current velocity.
The upper interaction score is given by
\begingroup
\setlength{\abovedisplayskip}{5pt}
\setlength{\belowdisplayskip}{5pt}
\setlength{\abovedisplayshortskip}{3pt}
\setlength{\belowdisplayshortskip}{3pt}
\begin{equation}
    \ell_i^{\mathrm{route}} =
    \min\!\left(
        1,\,
        \left[1-\frac{d_i}{d_{\mathrm{int}}}\right]_0^1
        +\delta_{\ell}
    \right).
    \label{eq:route_interaction}
\end{equation}
\endgroup
After fine-tuning, we use a prediction horizon of
$T_{\mathrm{int}}=5\,\mathrm{s}$, a distance scale of
$d_{\mathrm{int}}=0.5\,\mathrm{m}$, and an uncertainty allowance of
$\delta_{\ell}=0.15$ for live CRA.

\paragraph{Kinetic risk.}
Kinetic severity combines estimated mass with relative motion. We estimate the object's planar velocity $\mathbf{v}_i(t_L)$ from
the displacement of its LiDAR centroid in the world frame between
matched scans, divided by the elapsed time. We estimate the robot's
velocity $\mathbf{v}_{\mathrm{ego}}(t_L)$ from its pose history over a
short window ending at the current LiDAR timestamp $t_L$. The relative
speed is $v_{\mathrm{rel},i} =
\|\mathbf{v}_i(t_L)-\mathbf{v}_{\mathrm{ego}}(t_L)\|_2$. Kinetic severity is then computed as
\begingroup
\setlength{\abovedisplayskip}{4pt}
\setlength{\belowdisplayskip}{4pt}
\setlength{\abovedisplayshortskip}{2pt}
\setlength{\belowdisplayshortskip}{2pt}
\begin{equation}
    E_{i,\mathrm{kin}}^{+}
    =
    \frac{1}{2}m_i^{+}
    \left(v_{\mathrm{rel},i}^{+}\right)^2,
    \qquad
    s_{i,\mathrm{kin}}
    =
    \left[
        \frac{E_{i,\mathrm{kin}}^{+}}{E_{\mathrm{kin}}^{\star}}
    \right]_0^1,
    \label{eq:kinetic_severity}
\end{equation}
\endgroup
where $E_{\mathrm{kin}}^{\star}$ is a reference energy chosen to
represent high kinetic risk. At or above this energy, CRA assigns
the maximum kinetic severity, $s_{i,\mathrm{kin}}=1$. For the interaction score, 
we set $\ell_{i,\mathrm{kin}}=\ell_i^{\mathrm{route}}$, which
expresses the opportunity for interaction based on how close the
object's predicted path comes to the robot's planned route.

\paragraph{Potential risk.}
For object $i$, potential severity combines the estimated mass
with the vertical height difference
\(
    \Delta h_i =
    [h_i^{\mathrm{upper}}-h_i^{\mathrm{lower}}]_+ ,
\)
where $h_i^{\mathrm{upper}}$ and $h_i^{\mathrm{lower}}$ are the
heights of the upper and lower surfaces, respectively. We estimate
these heights from LiDAR observations subject to geometric checks.
The potential-energy severity is
\begingroup
\setlength{\abovedisplayskip}{4pt}
\setlength{\belowdisplayskip}{4pt}
\setlength{\abovedisplayshortskip}{2pt}
\setlength{\belowdisplayshortskip}{2pt}
\begin{equation}
    E_{i,\mathrm{pot}}^{+} = m_i^{+}g\,\Delta h_i^{+},
    \qquad
    s_{i,\mathrm{pot}} =
    \left[
        \frac{E_{i,\mathrm{pot}}^{+}}{E_{\mathrm{pot}}^{\star}}
    \right]_0^1 ,
    \label{eq:potential_severity}
\end{equation}
\endgroup
where $\Delta h_i^{+} =
[(h_i^{\mathrm{upper}})^{+}-(h_i^{\mathrm{lower}})^{-}]_+$.
The interaction score $\ell_{i,\mathrm{pot}}$ reflects how likely
the object is to affect the robot's planned route if it falls.
As with $E_{\mathrm{kin}}^{\star}$, we choose
$E_{\mathrm{pot}}^{\star}$ as a reference energy representing high
potential risk, which sets $s_{i,\mathrm{pot}}=1$ when
$E_{i,\mathrm{pot}}^{+}\geq E_{\mathrm{pot}}^{\star}$.

\paragraph{Contact-severity risk.}
We estimate contact severity from the contour of the object mask
and its associated exposed parts in the RGB image. We resample and smooth the outer boundary of the object mask. We
then estimate ``sharpness'' by corner-detector techniques~\cite{he2004curvature, zhang2007multi}.
At each boundary position $j$, we compute signed turning angles
$\alpha_{ij}^{(1)}, \alpha_{ij}^{(2)}, \alpha_{ij}^{(3)}$ over three
increasingly long stretches of the contour. Using the clipped ramp
$R_{a,b}(x)=[(x-a)/(b-a)]_0^1$, we compute the contour score as
\begingroup
\setlength{\abovedisplayskip}{5pt}
\setlength{\belowdisplayskip}{5pt}
\setlength{\abovedisplayshortskip}{3pt}
\setlength{\belowdisplayshortskip}{3pt}
\begin{equation}
\begin{split}
    z_i^{\mathrm{out}} = \max_j \Bigg\{&
        R_{\alpha_0,\alpha_1}\!\left(
            \min\big(\alpha_{ij}^{(1)},\alpha_{ij}^{(2)}\big)
        \right) \\
        &{}\cdot R_{\eta_0,\eta_1}\!\left(
            \frac{\alpha_{ij}^{(1)}}
                 {\max(\alpha_{ij}^{(3)},\epsilon)}
        \right)
    \Bigg\},
\end{split}
\label{eq:contour_contact}
\end{equation}
\endgroup
where $\alpha_0=25^\circ$, $\alpha_1=155^\circ$, $\eta_0=0.35$,
$\eta_1=0.75$, and $\epsilon>0$ prevents division by zero.
The two terms describe local convex-turn strength and the contour curvature across scales. 

To assess contact risk from exposed parts such as knife and scissor
blades for our specific restaurant experiments (Sec.~\ref{sec:experiment}),
we also fine-tune YOLO11n-seg on exposed-edge masks derived from
PACO~\cite{ramanathan2023paco}. We match each predicted part to its
object mask and compute a score based on its shape
descriptors~\cite{gonzalez2002digital}:
\begingroup
\setlength{\abovedisplayskip}{5pt}
\setlength{\belowdisplayskip}{5pt}
\setlength{\abovedisplayshortskip}{3pt}
\setlength{\belowdisplayshortskip}{3pt}
\begin{equation}
    z_i^{\mathrm{part}}
    = e_i\left(0.6\,\zeta_i+0.4\,\chi_i\right),
    \label{eq:part_contact}
\end{equation}
\endgroup
where $e_i$ is the fraction of the object mask covered by the part,
$\zeta_i$ measures how long and narrow the part is, and $\chi_i$ is
one minus its circularity. Finally, we set:
\begingroup
\setlength{\abovedisplayskip}{5pt}
\setlength{\belowdisplayskip}{5pt}
\setlength{\abovedisplayshortskip}{3pt}
\setlength{\belowdisplayshortskip}{3pt}
\begin{equation}
    s_{i,\mathrm{con}}
    = \max\!\left\{(z_i^{\mathrm{out}})^+,
                   (z_i^{\mathrm{part}})^+\right\},
    \qquad
    \ell_{i,\mathrm{con}} = \ell_i^{\mathrm{route}}.
\end{equation}
\endgroup

\paragraph{Thermal risk.}
For thermal risk, we use visible steam, smoke, or flames in RGB images as evidence that the associated object may be a heat source. We use pretrained SAM3~\cite{carion2026sam} with text prompts for
\textit{steam}, \textit{smoke}, \textit{fire}, and \textit{flame}
to obtain masks of the corresponding visible regions in the RGB image. However, in our \textit{Isaac Sim} experiments (Sec.~\ref{sec:experiment}), steam is rendered as animated white puffs. We therefore obtain steam masks by aligning the current RGB image with a steam-free reference image and detecting their differences. Finally, we associate each detected region with its potential source object, such as a pot or burner.

We estimate thermal severity from the mask area $A$, vertical
extent $V$, and persistence $P$. Area and extent are normalized
relative to the source region, and persistence is the fraction of
frames with an associated mask in the observation window. Using tuned weights for mask area, vertical extent, and persistence,
we compute the visual heat score as
\begingroup
\setlength{\abovedisplayskip}{5pt}
\setlength{\belowdisplayskip}{5pt}
\setlength{\abovedisplayshortskip}{3pt}
\setlength{\belowdisplayshortskip}{3pt}
\begin{equation}
    S^{+} = 0.40 A^{+} + 0.25 V^{+} + 0.35 P^{+}.
    \label{eq:thermal_visual_score}
\end{equation}
\endgroup
For masks obtained with SAM3, we multiply the visual heat score by
$0.65$ for \textit{steam} and \textit{smoke}, and by $1.0$ for
\textit{fire} and \textit{flame}. We take the thermal score $S^{+}$ as the thermal severity, setting
$s_{i,\mathrm{th}}=S^{+}$. In live navigation, we set
$\ell_{i,\mathrm{th}}=\ell_i^{\mathrm{route}}$.

\paragraph{Confidence adjustment and aggregation.}
For each risk factor, we multiply severity by interaction to obtain
a score $\overline{r}_{if}$. We choose an adjustable fallback value
$p_f$ to represent a typical level of risk for that factor. For the base CRA formulation, we set $p_f=0.25$. If there is no usable evidence to estimate the factor score, we set
$r_{if}=p_f$. If the unadjusted score
$\overline{r}_{if}$ is below $p_f$, lower confidence moves $r_{if}$
closer to $p_f$. If $\overline{r}_{if}$ is at or above $p_f$, we set
$r_{if}=\overline{r}_{if}$. We combine the four adjusted scores $r_{if}$ into
one risk score $\phi_i$ for subsequent safety filtering:
\begingroup
\setlength{\abovedisplayskip}{5pt}
\setlength{\belowdisplayskip}{5pt}
\setlength{\abovedisplayshortskip}{3pt}
\setlength{\belowdisplayshortskip}{3pt}
\begin{equation}
\begin{aligned}
    \overline{r}_{if} &= s_{if}\ell_{if},\\
    r_{if} &= \overline{r}_{if}
        + (1-c_{if})[p_f-\overline{r}_{if}]_+,\\
    \phi_i &= 1-\prod_{f\in\mathcal{F}}(1-r_{if}),
\end{aligned}
\label{eq:cra_aggregation}
\end{equation}
\endgroup
where $\mathcal{F}=\{\mathrm{kin},\mathrm{pot},\mathrm{con},\mathrm{th}\}$.
The product keeps $\phi_i$ between $0$ and $1$ and allows several
moderate factor scores to produce a higher overall risk.

\paragraph{Risk Updates with Retained Estimates.}

For static objects, we retain previously accepted mass estimates and
visual proxies for contact severity. When the camera mask is unavailable,
we reuse these values if the current LiDAR returns clearly match the
same object. Their confidence decreases with time since the original
observation, and we discard them once they expire. We combine the retained estimates with current observations to
recompute the factor scores and update the overall risk score $\phi_i$.

\subsection{Risk-Conditioned Guidance and Safety Filtering}
\label{sec:risk_conditioned_navigation}

Following the PSF and LGF construction in
Sec.~\ref{sec:psf_lgf}, we set the boundary parameters for each
object $i$ according to its CRA score:
\begingroup
\setlength{\abovedisplayskip}{5pt}
\setlength{\belowdisplayskip}{5pt}
\setlength{\abovedisplayshortskip}{3pt}
\setlength{\belowdisplayshortskip}{3pt}
\begin{equation}
    b_i = -(0.5 + 1.2\phi_i), \qquad \lambda_i = 1.25,
    \label{eq:risk_to_guidance}
\end{equation}
\endgroup
where $b_i$ is the normal boundary flux and $\lambda_i$ is the
tangential guidance strength. Higher risk increases the magnitude
of the normal flux, while the tangential strength remains fixed.

Using these boundary parameters, we construct the risk-conditioned
guidance field $\mathbf{v}_{\phi}$ and Poisson safety function
$h_{\phi}(\mathbf{q},t)$. To encourage smooth detours around obstacles, we include a soft cost for tracking a reference derived from the guidance field. The MPC safety layer then adjusts the nominal command sequence by solving

\begingroup
\setlength{\abovedisplayskip}{4pt}
\setlength{\belowdisplayskip}{4pt}
\setlength{\abovedisplayshortskip}{2pt}
\setlength{\belowdisplayshortskip}{2pt}
\begin{equation}
\begin{aligned}
&\mathbf{u}^{\star}_{0:N-1}
 = \operatorname*{arg\,min}_{\mathbf{u}_{0:N-1}}\;
   J_{\mathrm{nom}} + \beta_g J_{\mathrm{guid}} \\
&\text{s.t.}\;
\begin{aligned}[t]
&\mathbf{q}_{k+1} = \mathbf{F}(\mathbf{q}_k,\mathbf{u}_k), \\
&h_{\phi}(\mathbf{q}_{k+1},t_{k+1})
  \geq \rho\,h_{\phi}(\mathbf{q}_k,t_k), \\
&\mathbf{u}_k\in\mathcal{U}, \quad k=0,\ldots,N-1.
\end{aligned}
\end{aligned}
\label{eq:risk_conditioned_mpc}
\end{equation}
\endgroup
Here, $\mathbf{F}$ advances the model in
Eq.~\eqref{eq:navigation_dynamics} by one prediction step of duration
$\Delta t$. The initial state $\mathbf{q}_0$ is the current robot pose,
and $\rho=e^{-\gamma\Delta t}$, where $\gamma$ is the barrier gain. The cost $J_{\mathrm{nom}}$
penalizes deviation from the nominal commands. The cost
$J_{\mathrm{guid}}$, weighted by $\beta_g$, tracks a checked local
reference constructed from $\mathbf{v}_{\phi}$. The safety function
is evaluated at the look-ahead state associated with each predicted pose.

At execution, the analytical CBF makes the smallest weighted
adjustment to the first MPC command $\mathbf{u}^{\star}_0$ within
$\mathcal{U}$ that satisfies
\begingroup
\setlength{\abovedisplayskip}{5pt}
\setlength{\belowdisplayskip}{5pt}
\setlength{\abovedisplayshortskip}{3pt}
\setlength{\belowdisplayshortskip}{3pt}
\begin{equation}
    \mathbf{v}_{\phi}^{\top}\dot{\mathbf{p}}_{\ell}
    + \frac{\partial h_{\phi}}{\partial\psi}\,\omega
    + \sigma\frac{\partial h_{\phi}}{\partial t}
    \geq -\gamma h_{\phi},
    \label{eq:risk_conditioned_cbf}
\end{equation}
\endgroup
where $\dot{\mathbf{p}}_{\ell}$ is the planar velocity of the
look-ahead point induced by the command and $\sigma$ scales the
temporal derivative. All field quantities are evaluated at the
current look-ahead state.

\footnotetext{Video sources from Pexels:
\href{https://www.pexels.com/video/opening-pot-cover-and-steam-running-away-from-pot-11654923/}
{steaming pot},
\href{https://www.pexels.com/video/close-up-shot-of-fire-on-the-gas-stove-5969386/}
{gas burner}, and
\href{https://www.pexels.com/video/12935403/}
{wood fire}.}

\section{Experiments}
\label{sec:experiment}

\subsection{Experimental Setup}

We conduct the navigation experiments in NVIDIA \textit{Isaac Sim
6.0.1}~\cite{NVIDIA_Isaac_Sim} on a workstation equipped with an
NVIDIA GeForce RTX 5070 Ti. We adapt the Omniverse Restaurant Demo
Assets Pack for the restaurant
demonstration. \footnote{Web source: %
\href{https://docs.omniverse.nvidia.com/usd/latest/usd_content_samples/downloadable_packs.html\#restaurant-demo-assets-pack}
{NVIDIA Omniverse Restaurant Demo Assets Pack.}}

For the navigation experiments, we use a simulated Nova Carter differential-drive robot\footnote{Web source: %
\href{https://docs.isaacsim.omniverse.nvidia.com/6.0.1/assets/nova_carter_landing_page.html}
{NVIDIA Nova Carter documentation for Isaac Sim 6.0.1.}}
with a 32-channel Hesai PandarXT-32 3D LiDAR operating at 10~Hz
and a calibrated front-facing RGB camera.

\subsection{Continuous Risk Assessment Analysis}
\label{sec:cra_analysis}

We evaluate contact and thermal risk on RGB images and videos
(Fig.~\ref{fig:cra_factors}\subref{fig:cra_contact}
and~\subref{fig:cra_thermal}). In \textit{Isaac Sim}, we vary the
requested speed past a static obstacle to test kinetic risk
(Fig.~\ref{fig:cra_factors}\subref{fig:cra_kinetic}). For potential
risk, CRA uses the estimated depth of a shallow hole to guide the
robot around it (Fig.~\ref{fig:cra_factors}\subref{fig:cra_dropoff}).

We also evaluate CRA in a combined-risk scenario involving a human
carrying a steaming pot. CRA jointly considers the human's motion,
the potential risk of the elevated pot, and the thermal risk associated
with the visible steam (Fig.~\ref{fig:cra_combined}).

\subsection{Restaurant Navigation in Simulation}
\label{sec:restaurant_navigation}

We further evaluate CRA in a simulated restaurant with
live performance. The robot starts with a meal on its tray, delivers it to the left
dining area, returns to the service bar for a second meal, and
delivers it to the right dining area
(Fig.~\ref{fig:restaurant_navigation}). The complete
service sequence is shown in the video available on our
\href{https://isaac-restaurant-cra.cra-demos.workers.dev}{project webpage}.

\section{Conclusion}

Motivated by the idea of fuzzy safety, we presented Continuous Risk
Assessment (CRA) for safety-critical visual navigation. CRA uses RGB and LiDAR observations to estimate kinetic, potential,
contact-severity, and thermal risk. During navigation, we aggregate
these factors into a continuous risk score for each object or human
and use the resulting scores to shape the guidance and safety fields
for the MPC and analytical CBF layers. We evaluated CRA in \textit{Isaac Sim} on
a restaurant delivery mission. In future work, we plan to deploy CRA in real-world environments to evaluate its performance in continuous safety-critical situations across more complex social interaction scenarios.

\paragraph{LLM Usage Statement} The authors used ChatGPT for grammar and editing of the text, for polishing paper figures, for writing experiment code, and have reviewed and verified all AI-generated content for correctness.

\FloatBarrier

{\small
\bibliographystyle{ieeenat_fullname}
\bibliography{main}
}


\end{document}